\documentclass[letterpaper, 10 pt, conference]{ieeeconf}  
\usepackage[T1]{fontenc}

\IEEEoverridecommandlockouts                              

\usepackage{graphics} 
\usepackage{amsmath} 
\usepackage{amssymb}  
\usepackage{makecell}
\usepackage{graphicx}
\usepackage{wasysym}

\title{\LARGE \bf
A highly maneuverable flying squirrel drone \\ with controllable foldable wings
}

\author{Jun-Gill Kang*, Dohyeon Lee*, and Soohee Han 
\thanks{*Joint first authors.}
\thanks{The authors are all with Department of Convergence IT Engineering, Pohang University of Science and Technology (POSTECH), 37673 Pohang, South Korea.  \{jgkang1210, dohyeon, soohee.han\}@postech.ac.kr
}
}

\begin{document}

\maketitle
\thispagestyle{empty}
\pagestyle{empty}

\begin{abstract}

Typical drones with multi rotors are generally less maneuverable due to unidirectional thrust, which may be unfavorable to agile flight in very narrow and confined spaces. This paper suggests a new bio-inspired drone that is empowered with high maneuverability in a lightweight and easy-to-carry way. The proposed flying squirrel inspired drone has controllable foldable wings to cover a wider range of flight attitudes and provide more maneuverable flight capability with stable tracking performance. The wings of a drone are fabricated with silicone membranes and sophisticatedly controlled by reinforcement learning based on human-demonstrated data. Specially, such learning based wing control serves to capture even the complex aerodynamics that are often impossible to model mathematically. It is shown through experiment that the proposed flying squirrel drone intentionally induces aerodynamic drag and hence provides the desired additional repulsive force even under saturated mechanical thrust. This work is very meaningful in demonstrating the potential of biomimicry and machine learning for realizing an animal-like agile drone.

\end{abstract}


\section{INTRODUCTION}

Flying squirrel(Pteromyini) is one of the most acrobatic creatures in nature equipped with fast and efficient gliding capability. To reveal this creature’s outstanding control strategy, some attempts have been made to understand the complex aerodynamics occurring at its soft membrane wing in the challenging high angle of attack \cite{bishop2006relationship, ZhaoAerodynamic, paskins2007take, ando1993gliding, bahlman2013glide, norberg1985evolution, essner2003locomotion}. Furthermore, to mimic the flying squirrel’s gliding capability, biomimetics robots have been developed \cite{shin2019development, LiAirposture}.

Existing works focus on passive gliding capability by artificially providing additional power sources such as human power or external launcher devices. In nature, flying squirrels free fall from trees to convert its potential energy to kinetic energy thus gain sufficient speed for gliding and perching \cite{bishop2006relationship, LiAirposture, vernes2001gliding}. To make better use of a flying squirrel-shaped drone in the practical missions such as surveillance and reconnaissance purpose it would be very meaningful to develop an active control system for maneuvering its own body during flight phase with an onboard power system.

Conventional drones have internal control systems for surrounding rotors to maintain their body attitude \cite{bouabdallah2007full, 5649111, 9551712, 8892595}. However, for highly maneuvering flight, such control systems don't work properly due to limited speeds of rotors, leading to poor tracking performance \cite{faessler2016thrust}. This issue becomes serious for unexpected emergency cases and precise tracking tasks. As a specific example, an emergency stop is required when some objects suddenly appear in front. For this, additional repulsive force should be generated.

 Meticulously observing wing control strategies from flying squirrels and considering physical limitation of conventional drones, naturally motivate us to develop a new type of flying squirrel-shaped drone that has controllable foldable wings. If the wings of the proposed drone are folded, it is reduced to a conventional drone. The wings are unfolded when needed to gain more aerodynamic drag forces for sharp acceleration and stable precise tracking.

In this paper, in addition to the foldable wings mounted on drones, their control scheme is also covered for the first time. Actually, it is very hard to model complex nonlinear aerodynamics of flying squirrel’s elastic wings \cite{bishop2006relationship, ZhaoAerodynamic, shin2019development}. Accordingly, model-based wing control design is also not suitable. This paper suggests a data-driven approach using reinforcement learning (RL) from human demonstration data.

Specifically, a residual RL method is employed to extract the knowledge of human’s suboptimal trajectories and then enhance the training of RL agent \cite{johannink2019residual}.  Human demonstration data obtained by manual flight, is used to pretrain the baseline controller and train the auxiliary RL based policy for better performance.

The main contributions of this work can be summarized as follows:

\begin{itemize}
  \item \textit{World-first drone with controllable foldable wings} : 
Controllable foldable silicon membrane wings mimicking flying squirrels are fabricated to ensure proper thickness and elasticity. 
  \item \textit{Data-driven human-level wing control} :  Highly sophisticated wing control is efficiently designed with residual RL training and human demonstration data.
  \item \textit{Real experimental demonstration} : In a thrust-saturating trajectory tracking task, it is found that the developed silicone patagium governed by the learned wing control system can effectively produce aerodynamic drag in a desired direction.
\end{itemize}

The remainder of this paper is organized as follow. Section \ref{section2} describes how to implement foldable silicone wings with crank slider mechanism and construct their model. Section \ref{section3} describes how to control the proposed foldable wings of a drone by using an RL algorithm based on human demonstration data. Section \ref{section4} summarizes experimental data on real world. Finally, section \ref{section5} gives our conclusion and future research direction.

\begin{figure*}[t]
    \centering
    \includegraphics[width=\linewidth]{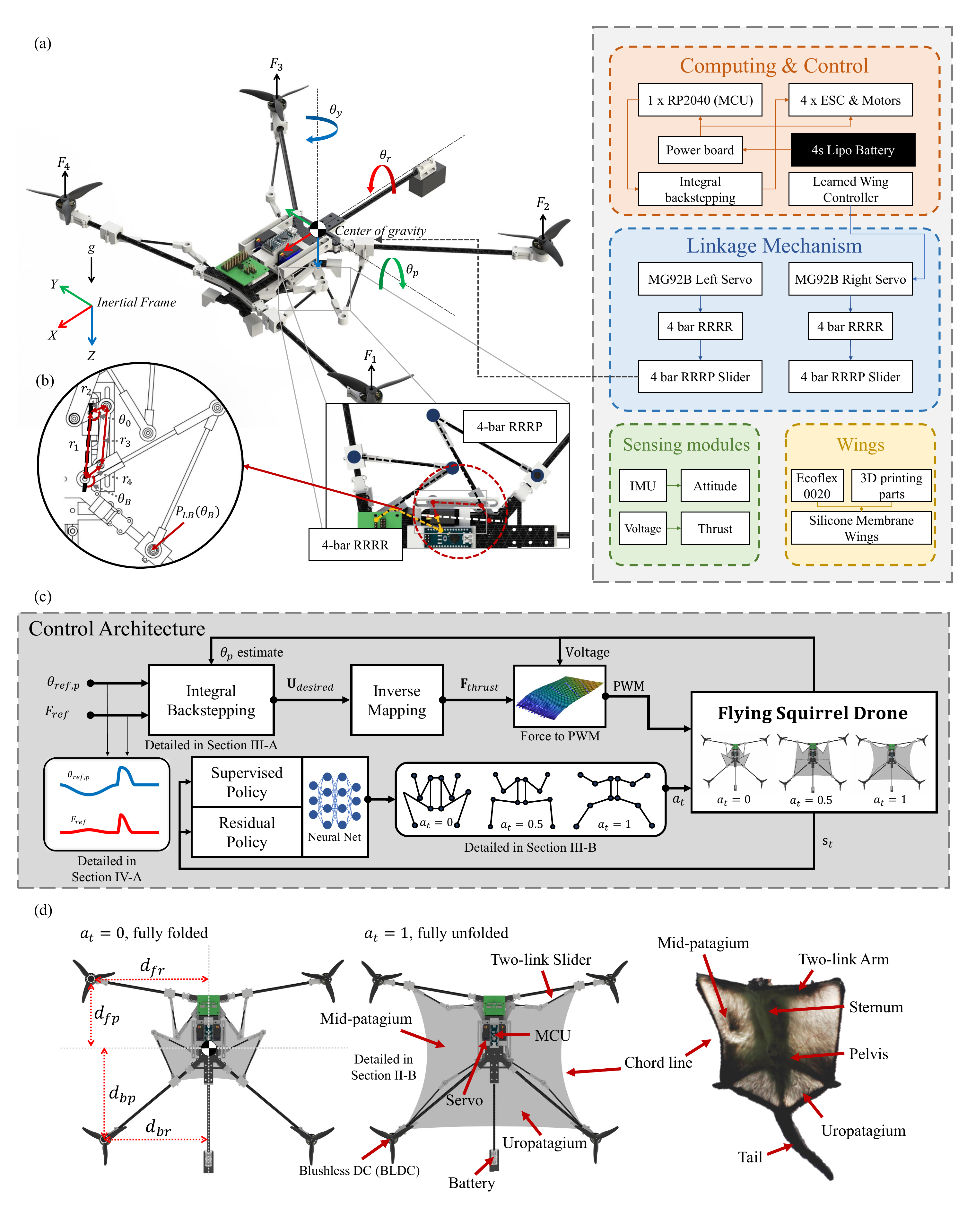}
    \caption{ Overview of the proposed flying squirrel drone : (a) Hardware components and related variables, (b) Enlarged view of the crank system, (c) Control block diagram, (d) Quadrotors with fully folded and unfolded wings, and a real flying squirrel with fully unfolded ones.}
    \label{structure_of_robot}
\end{figure*}

\section{Modelling of Robot \label{section2}}

In this section we introduce our flying squirrel quadrotor with detailed design process. Starting from hardware design we introduce silicon membrane fabrication process and finally wing folding mechanism.

\subsection{Hardware Design}

Table \ref{table_spec} summarize our hardware specification. The mass of the silicone wing is 46g which is less than 10\% of the weight of the robot. We developed flying squirrel drone using custom designed carbon plate frame with 3d printed mechanism parts. The main control loop for the system is entirely calculated on onboard micro controller unit(MCU) and an inertial measurement unit(IMU). We do not use commercial flight controller and design our own controller to handle special wing folding mechanism parts.

\begin{table}[ht]
\caption{Specifications of Robot}
\label{table_spec}
\begin{center}
\begin{tabular}{|c|c|}
\hline
\textbf{Parameters} & \textbf{Values} \\
\hline
\makecell{Dimensions } & \makecell{393.5~mm $\times$ 460.38~mm $\times$ 69.53~mm \\ (L $\times$ W $\times$ H) \\} \\
\hline
Mass & 0.635~kg (0.046~kg for silicone wings) \\
\hline
Moments of Inertia & \makecell{$I_{xx} = 0.005328198 (kg m^2)$ \\ $I_{yy} = 0.008677313 (kg m^2)$ \\ $I_{zz} = 0.013775190(kg m^2)$} \\
\hline
MCU & 1 $\times$ RP2040 \\
\hline
Power source & 1 $\times$ 14.8~V, 75C LiPo battery \\
\hline
Motors & \makecell{4 $\times$ T-motor 2004(3000 KV) \\ 2 $\times$ MG92B servo} \\
\hline
Sensors & \makecell{1 $\times$ IMU, 4 $\times$ ESC} \\
\hline
\end{tabular}
\end{center}
\end{table}

\subsection{Silicone Membrane Wings Fabrication}

We fabricated silicone wings that mimic the flying squirrel's patagium to induce aerodynamic drag. To ensure sufficient elasticity we choose silicone rubber to be our wings material. Silicone rubber is elastomer containing carbon, oxygen and silicone. Thanks to twisted covalent bond inside elastomer it has very low young's modulus and great flexibility.

\begin{figure}[t]
    \centering
    \includegraphics[width=7.5cm]{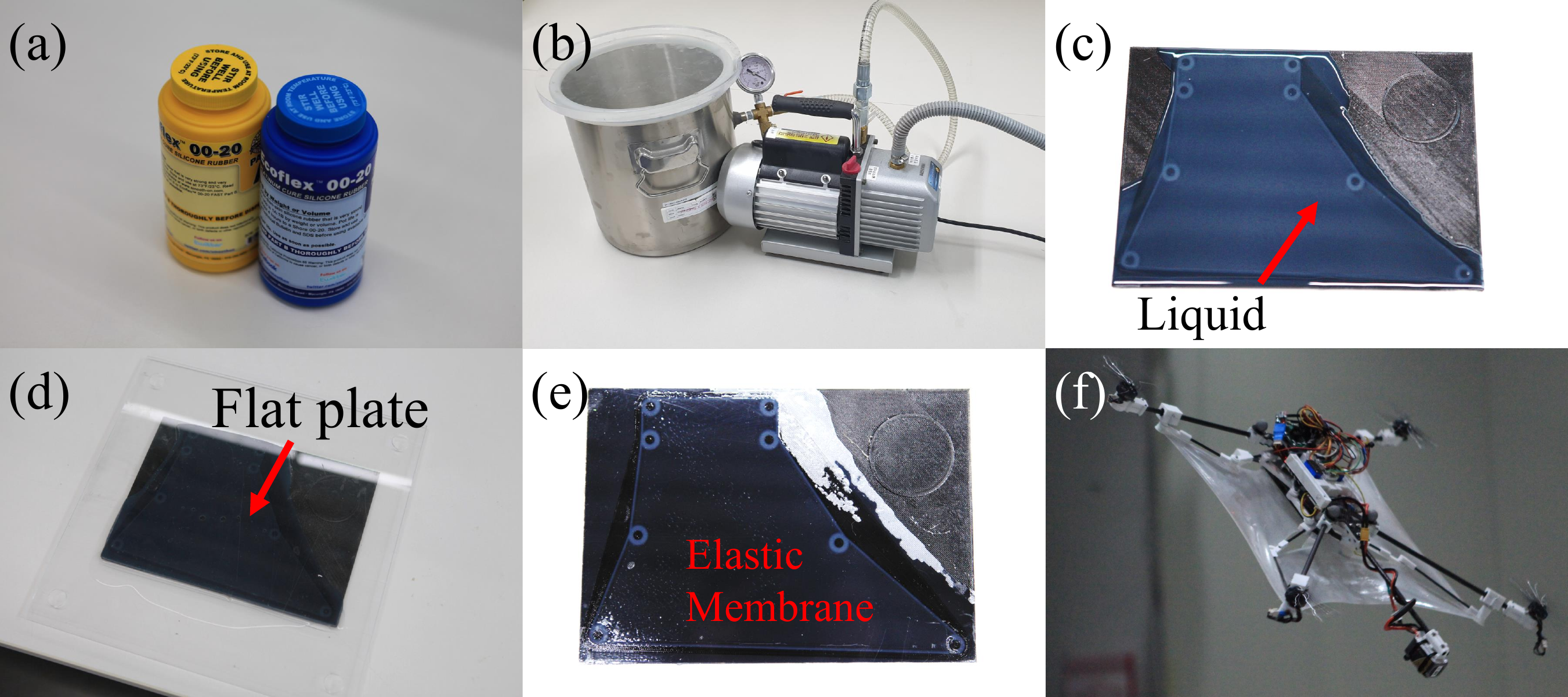}
    \caption{Fabrication process of Silicone Wings: (a) Ecoflex 0020 Silicone (b) Vacuum pump (c) 3d Printed mold (d) Press with Flat Plate (e) Obtained wings (f) Deployed on hardware.}
    \label{silicone_wing}
\end{figure}

Fig. \ref{silicone_wing}. shows our fabrication steps of silicone wings. We use Ecoflex 0020 from Smooth On, Inc. After mixing the silicone we used vacuum pump to remove all the bubble inside the silicon because during hardening of liquid silicone the bubble will make small hole inside silicone wings and let the wings vulnerable to shear stress. Next we poured our silicone in the 0.3mm thickness 3d printed mold and covered with flat plate to firmly press using sufficient weights. After waiting for 10 hours we can get elastic membranes which can be deployed to hardware directly.

Table \ref{table_silicone} shows our Ecoflex 0020 have similar tensile strength with respect to bio-inspired bat robot \cite{ramezani2017biomimetic}. If the tensile strength is too high then the silicone would have very large cohesive force thus hard to fabricated thinly. Also we note that this table is provided by manufacturer and it may have different characteristic since we made our silicone wings very thinner than normal usage.

\begin{table}[ht]
\caption{Specifications of Silicone}
\label{table_silicone}
\begin{center}
\begin{tabular}{|c|c|c|c|c|c|}
\hline
\textbf{Silicone} & \makecell{Mixed \\
viscosity} & \makecell{Tensile \\ 
strength} & \makecell{100\% \\
modulus} & \makecell{Elongation \\
at break \%} \\
\hline
\makecell{Ecoflex 0020 \\
(Ours)} & \makecell{3,000cps}  & \makecell{160psi}
 & \makecell{8psi}
  & \makecell{845\%}
   \\
\hline
\makecell{Ecoflex 0030} & \makecell{3,000cps}  & \makecell{200psi}
 & \makecell{10psi}
  & \makecell{900\%}
   \\
\hline
\makecell{DragonSkil 0010 \\ \cite{shin2019development}} & \makecell{23,000cps}  & \makecell{475psi}
 & \makecell{22psi}
  & \makecell{1000\%}
   \\
\hline
\makecell{Bat Bot \cite{ramezani2017biomimetic}} & \makecell{N.A.}  & \makecell{117.48psi}
 & \makecell{N.A.}
  & \makecell{439.27\%}
   \\
\hline
\end{tabular}
\end{center}
\end{table}

\subsection{Wing Folding Crank Slider Mechanism}

To mimic the folding leg structure of flying squirrels, we designed a 1 degree-of-freedom crank slider mechanism. The left and right wings are each controlled by a servo motor and their surface area is adjusted by the motion of sliders located above and below them. To achieve linear motion of the sliders, we combined two 4 bar linkage mechanisms: a 4-bar RRRR(4 rotation joint) linkage mechanism directly connected to a servo motor and another 4-bar RRRP(3 rotation 1 planar joint) mechanism that slide through quadrotor arm, resulting in a 6-bar linkage mechanism similar in structure. This linkage structure allowed us to adjust the range of motion of the sliders within limited hardware conditions. To model the aerodynamics of the flying squirrel drone, we needed to calculate the wing surface area according to the servo angles. There are a total of 12 points that affect the wing membrane area, and the locations of points that are not fixed points among them are calculated using the Freudenstein equation \cite{ghosal2010freudenstein}.

\begin{equation}\label{Freudenstein eqn1}
    R_1 cos\theta_B - R_2 cos\theta_0 + R_3 = cos(\theta_0-\theta_B)
\end{equation}

where

\begin{equation}\label{Freudenstein eqn2}
    R_1 = \frac{r_1}{r_2},  R_2 = \frac{r_1}{r_4}, R_3 = \frac{r_2^2+r_4^2+r_1^2-r_3^2}{2r_2r_4}
\end{equation}

$r_1$, $r_2$, $r_3$, $r_4$, $\theta_0$, $\theta_B$ and $P_{LB}$ are shown in (b) of Fig. \ref{structure_of_robot}. 
The relationship between $\theta_0$ and $\theta_B$ connected by a 4-bar link is obtained through equation \ref{Freudenstein eqn1}. and based on this, the position of the slider $P_{LB}$ could be expressed as a function of the angle of the servo. The positions of the front slider and joint points were expressed as a function of $\theta_0$ in the same way.

\begin{figure}[t]
    \centering
    \includegraphics[width=\linewidth]{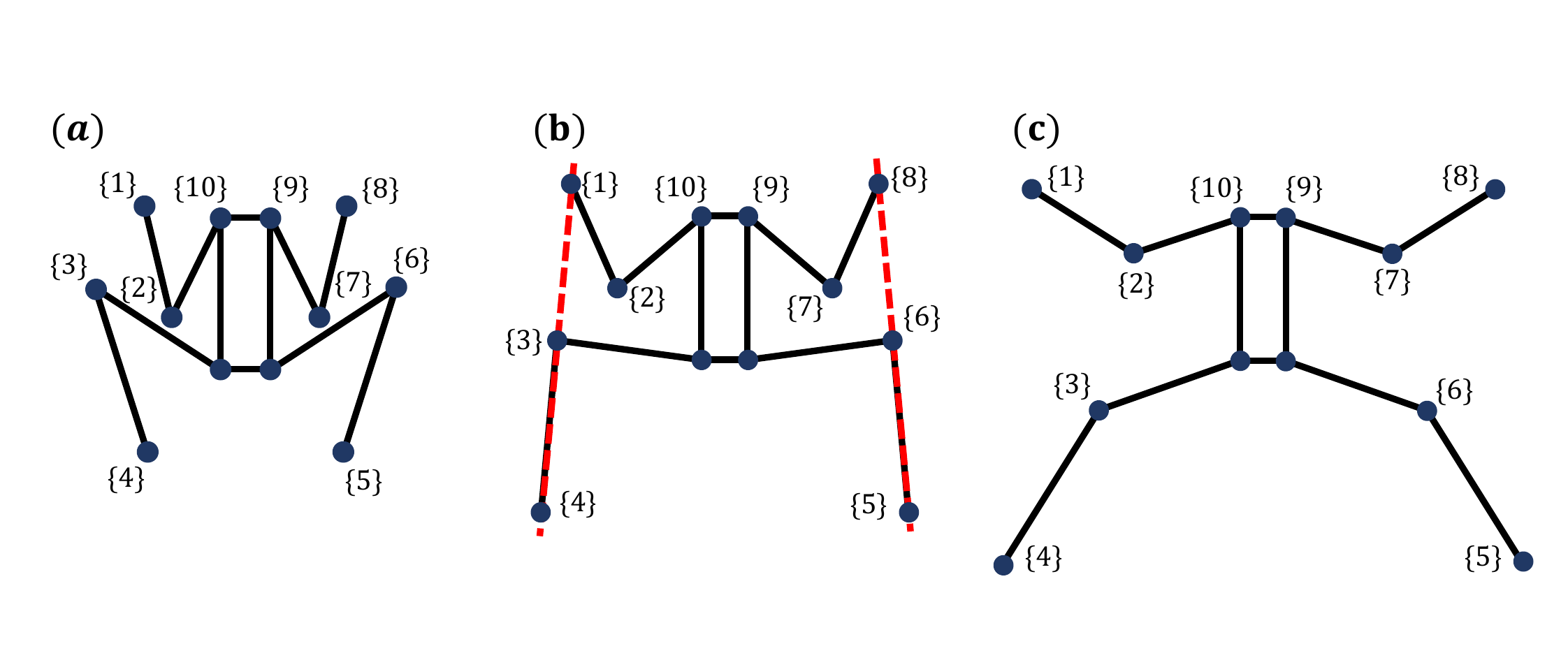}
    \caption{Wing area states.}
    \label{wingmechanism-area}
\end{figure}

The wing area was calculated using the Shoelace Formula from equation \ref{Shoelace_Formula}. The types of polygons made by the points that make up the area vary depending on the degree of wing folding. Based on (b) of Fig. \ref{wingmechanism-area} , where $\{1\}$, $\{3\}$, and $\{4\}$ points are placed in one straight line, it becomes a decagon when folded further and an octagon when unfolded.
\begin{equation}\label{Shoelace_Formula}
    A = \frac{1}{2}\left|\sum_{i=1}^{n-1}x_iy_{i+1}+x_ny_1-\sum_{i=1}^{n-1}x_{i+1}y_i-x_1y_n \right|
\end{equation}

Simplified aerodynamics using flat wing theory from \cite{perching2009} in equation \ref{Aerodynamics} with additional randomization to lift and drag coefficients was added for force calculation in simulation.

\begin{equation}\label{Aerodynamics}
    C_{L} = 2 \sin({\theta_p}) \cos({\theta_p}),  C_{D} = 2 \sin^2({\theta_p})
\end{equation}

With the lift and drag coefficients and area from equation \ref{Shoelace_Formula} we can obtain aerodynamic lift and drag forces as follows.

\begin{equation}\label{Aerodynamics_forces}
    F_{L} = \frac{1}{2} \rho_{air} C_{L} A v^2,  F_{D} = \frac{1}{2} \rho_{air} C_{D} A v^2
\end{equation}

Where air density $\rho_{air}=1.23[kg/m^3]$, $C_{L}$, $C_{D}$ from equation \ref{Aerodynamics}, and velocity of robot $v$ in aerodynamic reference frame.

\section{Controller Design \label{section3}}

We separate our quadrotor controller from wing controller for modular design. Quadrotor controller is based on Integral backstepping controller commonly used in literature. We designed our controller dynamics based on folded wing state which we call nominal quadrotor mode.

\subsection{Integral Backstepping Controller}\label{section3-A} 
The dynamic equations of a quadrotor contain non-linear terms such as gyroscopic and Coriolis forces. Thus we applied an Integral backstepping controller based on Lyapunov stability to our drone hardware, resulting in the pitch direction torque input of the drone as shown in equation \ref{Integral Backstepping}.

\begin{equation}
\begin{aligned}\label{Integral Backstepping}
    U_{p} = I_{yy}[(1-k_1^2+k_2)e_{\dot{\theta}_p}+(k_1+k_3)e_{\dot{\theta}_p}-\\
    k_1k_2\int e_{\theta_p}dt+\ddot{\theta}_{p}^{d}-\dot{\theta}_r\dot{\theta}_y\frac{I_{zz}-I_{xx}}{I_{yy}}+\frac{J_R}{I_{yy}}\dot{\theta}_p\Omega] 
\end{aligned}
\end{equation}
With error of pitch $e_{\theta_p} = \theta_{p}^{d}-\theta_p$ where $\theta^{d}_{p}$ is desired pitch and $\theta_{p}$ is current pitch. Same for pitch angular velocity error $e_{\dot{\theta}_p}$ with desired and current angular velocity $w_{r}^{d}$, $\dot{\theta}_p$. $J_R$ represents the rotor inertia and $\Omega$ represents the propeller angular velocity.
Roll and yaw inputs, $U_r$ and $U_y$ can also be obtained through a similar process. See \cite{bouabdallah2007full} \cite{5649111} for more information.

The derived torque input is mapped to the thrust of each motor using the following forward and inverse thrust to torque mapping by equations \ref{Integral Backstepping_translate_1} and \ref{Integral Backstepping_translate_2}. The thrusts F1, F2, F3 and F4 are shown in (a) of Fig. \ref{structure_of_robot}.
\begin{equation}
\begin{aligned}\label{Integral Backstepping_translate_1}
    &\textbf{U}=\textbf{TF}\\
    &\textbf{F} = \textbf{T}^{-1}\textbf{U} 
\end{aligned}
\end{equation}
With: 
\begin{equation}
\begin{aligned}\label{Integral Backstepping_translate_2}
\scriptstyle
    \textbf{U} = \left [ \begin{matrix} 
    U_{F} \\
    U_{r} \\ 
    U_{p} \\ 
    U_{y} \end{matrix}
    \right ], 
    \textbf{T} = \left [ \begin{matrix} 
    1&1&1&1 \\
    d_{fr}&d_{br}&-d_{br}&-d_{fr} \\ 
    d_{fp}&-d_{bp}&-d_{bp}&d_{fr} \\ 
    -ct_1&ct_2&-ct_1&ct_2\end{matrix}
    \right ], 
    \textbf{F} = \left [ \begin{matrix} 
    F_1 \\
    F_2 \\ 
    F_3 \\ 
    F_4 \end{matrix}
    \right ]
\end{aligned}
\end{equation}

Here $U_F$ is reference force that controls the altitude. We set $U_F$ to always satisfy equilibrium in gravitational direction. $ct_1$ and $ct_2$ are the ratio of torque to thrust produced by the propeller when the rotor rotates in the clockwise and counterclockwise directions, respectively. $d_{fr}$, $d_{br}$, $d_{fp}$, $d_{bp}$ is indicated in (d) of Fig. \ref{structure_of_robot}. When the drone spreads its wings, the center of gravity of the silicon membrane retreat by 27.6mm which is successfully handled by our feedback controller combined with learned policy.



\subsection{Learned Wing Controller}\label{section3-B}

\begin{figure}[b]
  \includegraphics[width=\linewidth]{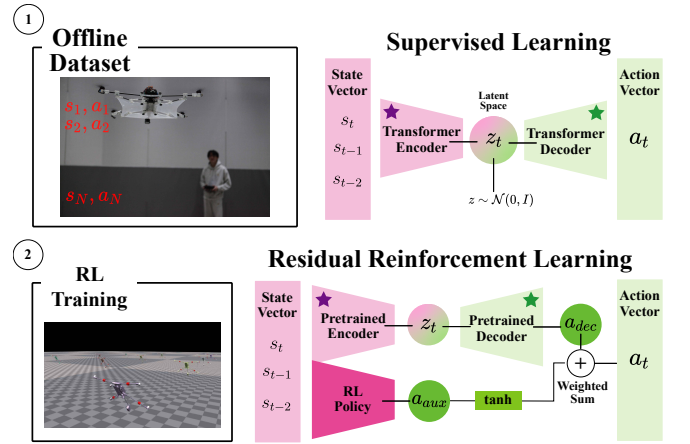}
  \caption{We first obtain offline dataset to be pretrained using supervised learning from human demonstrated data. Next we train with simulator using pretrained policy as base controller and learn the auxiliary policy for stable and fast learning performance. Star marked encoder, decoder is fixed during RL training.}
  \label{rl_train}
\end{figure}

Finding optimal wing controller for flying squirrel drone is extremely hard due to complex aerodynamics. But often human can learn and control the robot system without knowledge of full dynamics of system using intution based control. In this work we obtained human demonstrated dataset to capture the dynamics and suboptimal control strategy from human.

First we pretrained our policy using supervised learning using Transformer-Variational Auto Encoder(VAE) \cite{petrovich2021action} which is known to have great ability of learning time series data. We collected human demonstration for 3 times for each trajectories of 2.0s, 2.3s and 2.6s thus total 9 trajectories. For each trajectories all the reference trajectory was built from Section \ref{section4-A} with minimum pitch angle -40$^{\circ}$ and maximum pitch angle as 70$^{\circ}$. To handle overfit we augmented 9 trajectories data with Gaussian noise to obtain 100 augmented data for each trajectory thus total 900 samples. For positional states we added noise from normal distribution $0.001 \mathcal{N}(0, I)$ which is about variance of 1cm and for attitude of drone we added noise from $0.01745 \mathcal{N}(0, I)$ which is about variance of 1 degree. We do not add any noise to human guided action. After mixing the noise we numerically differentiated the data to obtain linear and angular velocity and use as states of the drone.



After obtaining data we augment this state action pair by adding random Gaussian noises to states and obtained total 909 trajectories. Due to the space limit of our motion capture system we were only able to obtain full state action pair trajectories for 2.0s and 2.3s trajectory data. For 2.6s trajectories we only obtained the data after acceleration phase.

Input state $s_t$ of our dataset contains linear velocity $v_x, v_y, v_z$, angular velocity $w_x, w_y, w_z$, gravitational direction $g_x, g_y, g_z$, reference pitch theta and reference altitude force $\theta_{ref,p},F_{ref}$, and previous action from human $a_{t-1}$, total 12 states. Target action of our dataset is action from human $a_t$.

For more stable training we let the neural network to see the three states including past two step's states $s_{t-2},s_{t-1},s_{t}$ to predict the current action $a_t$. The training objective was set to correctly match the action with given 3 states with additional Kullback-Leibler(KL) divergence regularization in latent space to preventing overfitting.

The action was recorded by measuring the time when the human tried to unfold the wings and fold the wings to gain maximum aerodynamic drag forces. The action $a_t$ of the wings are parameterized from 0 to 1 so that 0 denotes folded state and 1 denotes the unfolded state. For simplicity we set the left and right wing to move the same.


For RL training detail we used isaac gym simulation \cite{makoviychuk2021isaac} with 4096 parallel environments with 240Hz same with the state action pair dataset which is obtained by 240Hz motion capture data. Our state of the RL agent is same as dataset's 12 states and action is single number from 0 to 1 that maps to unfolding state and folding state of wing mechanism. We implemented same quadrotor controller inside the simulator to match the dynamics. 

We solved the Markov decision process (MDP) by a tuple $\{\mathcal{S},\mathcal{A},\mathcal{T},R,\rho,\gamma\}$ of states, actions, transition probability reward, initial state distribution, and discount factor. Using supervised pretraining step we obtained human demonstrated policy $\pi_{\phi_{h}}(a|s)$ with the parameter $\phi_{h}$ to follow the human demonstration. But this policy is highly suboptimal so we use residual reinforcement learning to train $\pi_{\phi_{r}}(a|s)$ using Proximal Policy Optimization(PPO) \cite{schulman2017proximal} to maximize discounted reward.

The action for the simulation is calculated as follows. First we obtained $s_t$ from simulation. And we passed states to both $\pi_{\phi_{h}}(a|s)$ and $\pi_{\phi_{r}}(a|s)$ to get action $a_{dec}, a_{aux}$. The output of the $\pi_{\phi_{r}}(a|s)$ is activated by tanh activation function and multiply with fixed gain to work from fixed range centered at zero. Next we added both action and use this action to be current state's action. Final action $a_t$ can be expressed as follows $a_t = a_{dec} + c_{aux} \tanh({a_{aux}})$. For $c_{aux}$ we found that 0.75 was sufficient for 2.0s and 2.3s trajectories, and 1.0 for 2.6s trajectories.


For training we maximize 3 reward terms. First reward is to maximize the velocity right after acceleration phase thus maximizing $v_B$ from Fig. \ref{reference_traj_example}. Second term was set to minimize the velocity at terminal state with respect to maximum velocity. This can be written as follows $|v_B - v_{E}|$. Note that these two terms are adversarial so we need to select appropriate gain to prevent collapsing of training. Final term was to penalize action rate thus minimizing $|a_t - a_{t-1}|$ to ensure smooth action commands. The reward coefficients was fixed to be 5, 10, -0.275 for each terms across all tasks.

The detailed hyperparameters for training is described on appendix.

\section{Experiments \label{section4}}

\begin{figure}[b]
    \centering
    \includegraphics[width=7.0cm]{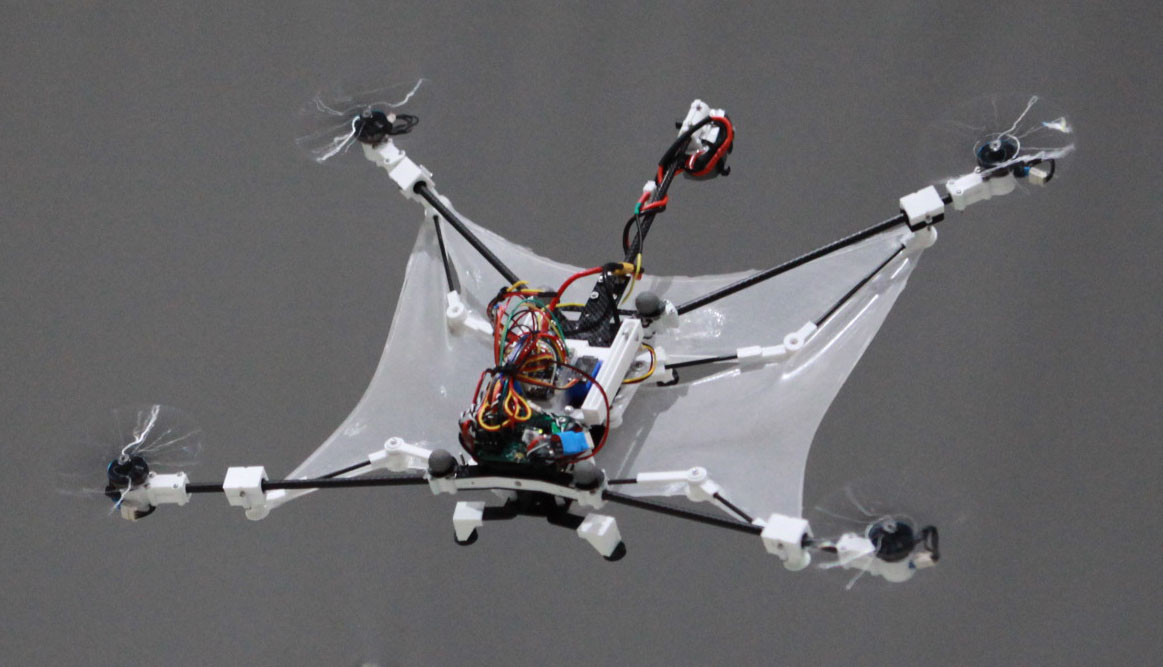}
    \caption{Developed flying squirrel drone used in experiments.}
    \label{developed_fs}
\end{figure}

\begin{figure*}
  \centering
  \includegraphics[width=17cm]{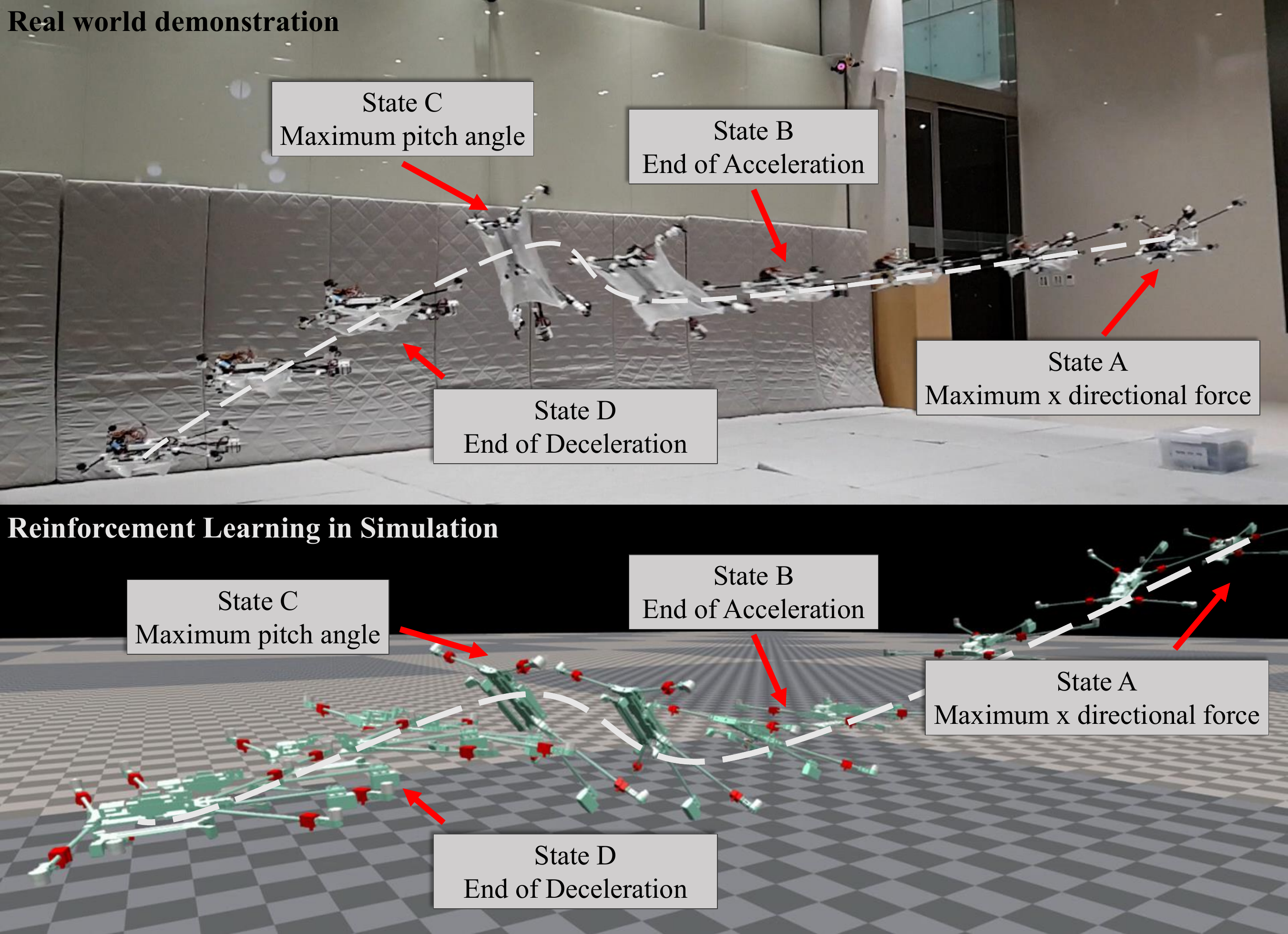}
  \caption{Flight experiments in real world with fabricated silicone wing and control algorithm. We can clearly see the distinct states A,B,C and D with transition of quadrotor to flying squirrel mode that was generated in section \ref{section4-A}. Additionally we also train the wing controller in simulation which is comparable to the real world demonstration.}
  \label{transition}
\end{figure*}

Our main objective of this paper is to show that the developed flying squirrel drone system can be effective even when the motors have been saturated due to thrust limit. To prove our objective we set some fixed trajectories with different total times. Each trajectories have distinctive acceleration and deceleration phases. Because of sharp acceleration commands the motors saturated during experiments.

\subsection{Reference Trajectory Following} \label{section4-A}

\begin{figure}[t]
    \centering
    \includegraphics[width=\linewidth]{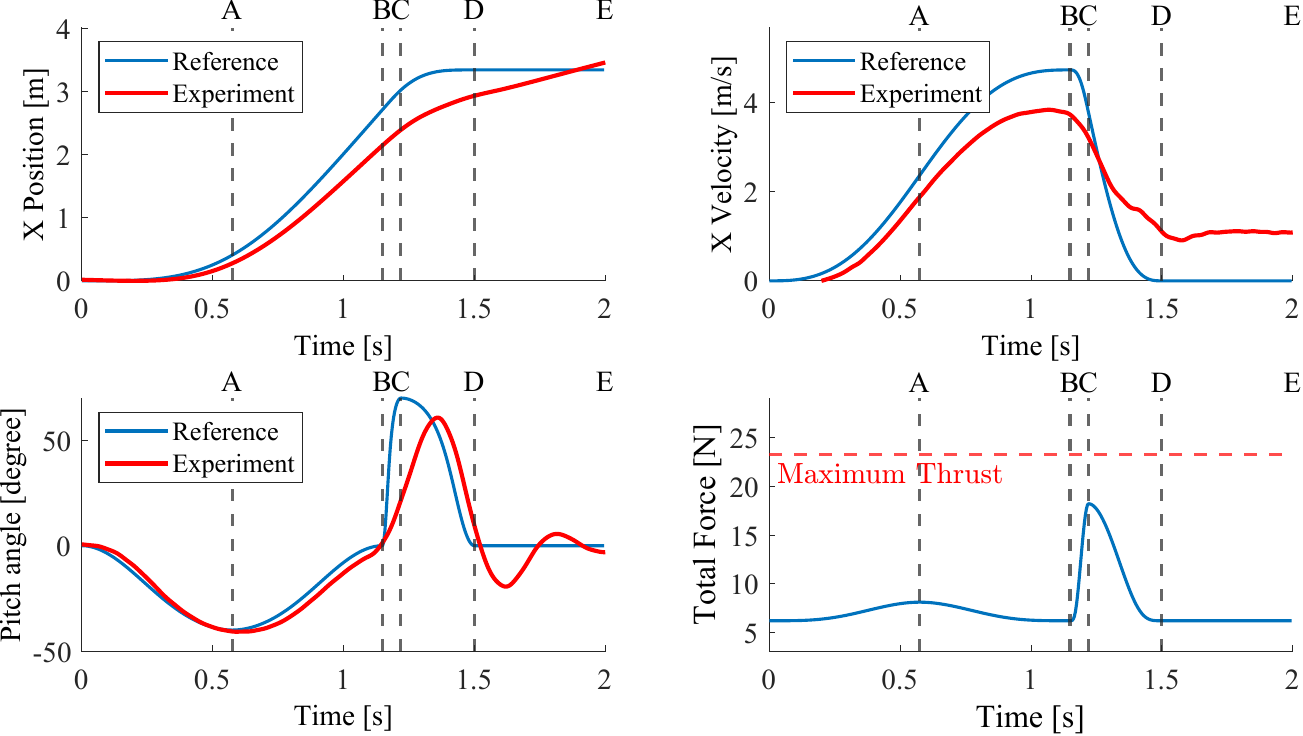}
    \caption{Example of reference trajectory when desired minimum pitch angle is -40 degree and maximum pitch angle is 70 degree. Initial and terminal velocity is set to be zero. Blue line represents the desired simulated results and red line represents the experimental data. We can clearly see that quadrotor couldn't follow the sudden deceleration commands after state B.}
    \label{reference_traj_example}
\end{figure}

In this study we compensate the gravitational force with proper y directional force to always satisfy equilibrium : $F_y = m g$ thus set the $v_y$ as zero. Also we assume that quadrotor always aligned with the command angle thus satisfying following equation : $F_x = F_y \tan(\theta_{p})$.

Now by only modifying x directional force $F_x$ we can shape our x directional trajectories. To create smooth and flexible reference trajectories we select $3^{rd}$-order B\'ezier curve to be our reference angle and force similar as \cite{park2015variable}. We first set minimum and maximum angle we want to achieve during operations. After setting the minimum and maximum angle we can calculate the minimum and maximum x directional force using equation : $F_x = F_y \tan(\theta_{p})$.

Additionally we set the initial and final $v_x$ to be zero by achieving equation \ref{x_force_matching} which is condition for x directional momentum conservation.

\begin{equation} \label{x_force_matching}
\int F_x dt = 0
\end{equation}

By setting the initial and final $v_x$ as zero we were able to easily compare experiments data by comparing the terminal velocity in fixed duration. Fig. \ref{reference_traj_example}. shows example of our trajectory. We divided our trajectory in 4 phases. In phase 1 starting from initial state to state A in figure our quadrotor achieves maximum angle thus creates maximum $F_x$. At phase 2 from A to B quadrotor finishes the acceleration phase. Now at phase 3 from B to C the quadrotor starts decelerate and achieve maximum deceleration force at state C. Finally from C to D the quadrotor finishes deceleration which is phase 4. We set rest commands for additional 0.5s to compare the converged terminal velocity after acceleration.

Because these force trajectories will always satisfy equation \ref{x_force_matching} the desired terminal $v_x$ is always set to be zero. Also we set the time ratio between phase 3 and phase 4 to be 1:4 to give quadrotor sharp acceleration commands. We can clearly see in Fig. \ref{reference_traj_example}. that the quadrotor was not able to follow the sharp acceleration commands during deceleration phase and couldn't set the terminal velocity to zero.

\subsection{Effectiveness of Human Demonstration}\label{section4-B}

We compared our learned controller from human demonstration with naively trained controller and human demonstrated data. We set two main criteria for wing controller. First the wing controller should not be unfolded during acceleration phase to gain maximum speed before deceleration phase. Second the wing controller should minimize the terminal velocity effectively. In this section we compare our controller with first criteria.

We compared three wing controllers naive RL, human, and residual RL. For naive RL we trained with all the same hyper parameter with residual RL except that we don't use the pretrain policy for training and we use sigmoid activation rather than tanh activation for final output. Human and residual policy was obtained following the steps in section \ref{section3-B}.

\begin{figure}[t]
  \centering
  \includegraphics[width=\linewidth]{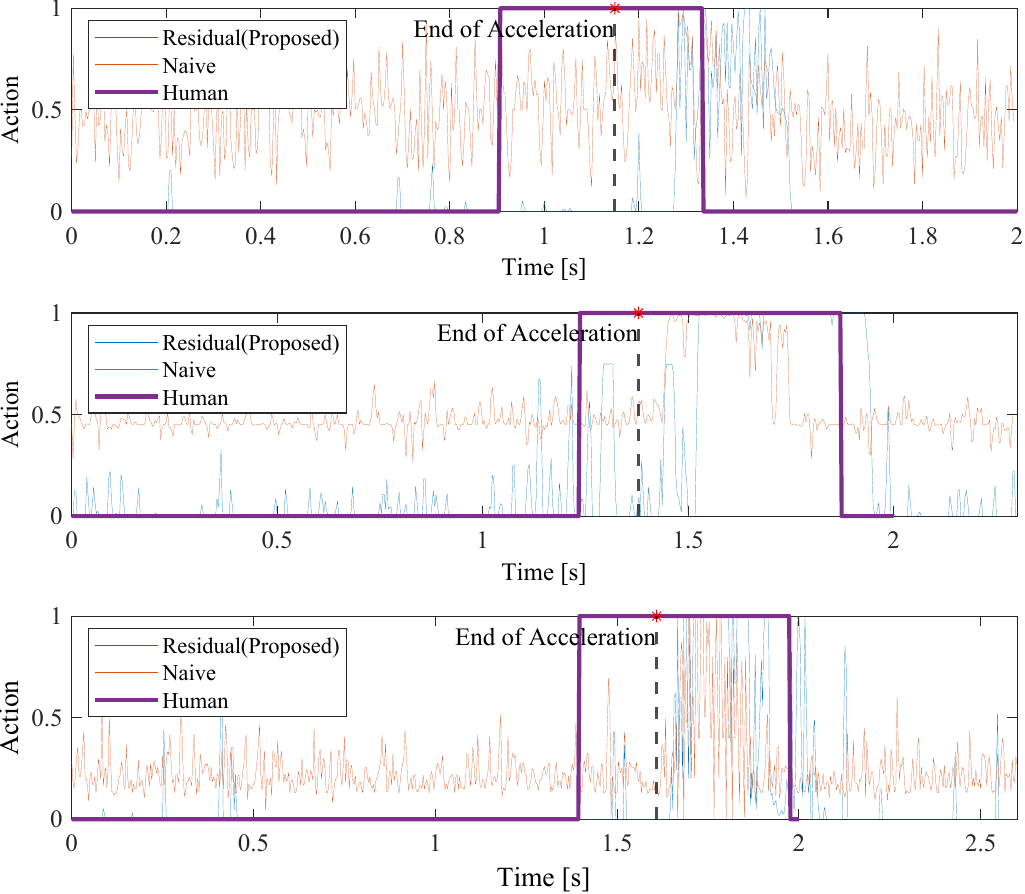}
  \caption{Comparison between proposed residual RL controller, naive RL controller, and human demonstrated data(averaged over 3 experiments); Top : 2s trajectory, Middle : 2.3s trajectory, Bottom : 2.6s trajectory.}
  \label{compare_action}
\end{figure}

In Fig. \ref{compare_action}. we can clearly see that naively trained RL policy converged to highly suboptimal policy. For 2s trajectories naive RL policy doesn't find out when to fold or unfold the wings and always set the wings at the middle of the two states. For 2.3s and 2.6s trajectories naive RL policy does find similar folding and unfolding timing with residual policy but doesn't fold the wings at acceleration phase thus interrupting the acceleration.

For human demonstrated data we found out that even though human pilot was trying to unfold the wing after the acceleration phase to achieve maximum speed at the beginning of the deceleration phase, all of the 9 trajectories unfolded the wings before acceleration phase has been ended.

Only residual RL trained policy was able to understand the problem objective and set action command near 0 (folded state) during acceleration phase and unfold the wings after end of the acceleration phase. At 2.3s trajectory in Fig. \ref{compare_action} the action exceeded over 0.5 before end of acceleration for very short time. But this fast action sequence can't be followed by our servo motor control bandwidth thus does not affect control performance.


This results show that even though our pretrained policy from human demonstrated data was not able to unfold the wings after acceleration, our auxiliary policy successfully help overall policy to learn optimal policy.

\subsection{Effectiveness of Foldable Wings}\label{section4-C}

\begin{table}[ht]
\caption{Decreased Velocity During deceleration}
\label{table_decrease}
\begin{center}
\begin{tabular}{|c|c|c|}
\hline
\textbf{Total Time} & \makecell{Residual(Proposed)} & \makecell{Nominal} \\
\hline
\makecell{2s} & \makecell{2.6981[m/s]}  & \makecell{2.6564[m/s]} \\
\hline
\makecell{2.3s} & \makecell{3.8171[m/s]}  & \makecell{3.7581[m/s]} \\
\hline
\makecell{2.6s} & \makecell{4.9526[m/s]}  & \makecell{4.8040[m/s]} \\
\hline
\end{tabular}
\end{center}
\end{table}

\begin{figure}[t]
    \centering
    \includegraphics[width=\linewidth]{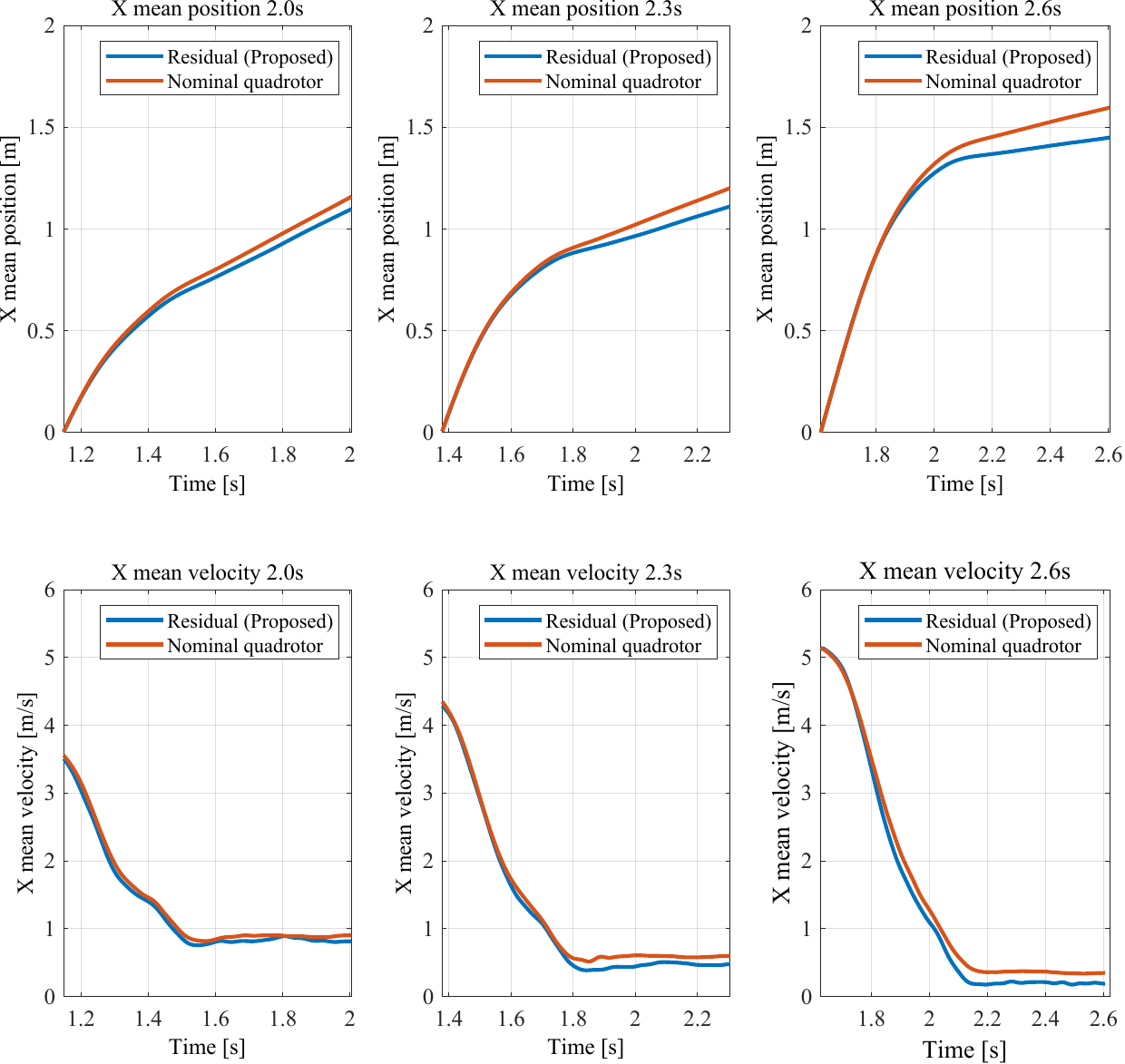}
    \caption{2.0s, 2.3s, 2.6s running time from left.}
    \label{residual_vs_nominal}
\end{figure}

Finally we compared our developed flying squirrel quadrotor with same quadrotor with fixed wings which will be denoted by nominal quadrotor and fold the wings always. For the residual RL controller we assumed that the states that the quadrotor encounter will be similar across experiments since we used fixed trajectories profile. With this assumption we can obtain the wing control sequence from simulation and replay in the real world experiments. Future work will include online wing controller.

Table \ref{table_decrease} summarize our results. By increasing the total time of trajectory the proposed hardware system with residual RL controller effectively decrease the velocity compare to nominal controller. The more we increase the total time of trajectory the more speed that the quadrotor will achieve and saturated more easily, but our system can effectively handle such problem by inducing drag forces.

Fig. \ref{residual_vs_nominal}. is graph showing the position and velocity of quadrotor during deceleration phase. We can see that our hardware system with proposed residual controller clearly decreased the terminal velocity.

Even though we were not able to conduct experiment for total trajectory time longer than 2.6s due to space limit of motion capture system, we can conclude our overall system effectively decreased the velocity compared to nominal quadrotor.

\section{CONCLUSIONS AND FUTUREWORKS \label{section5}}

In this work we present novel quadrotor with flying squirrel wing mechanism to breaks the limit of motors thrust saturation. We summarize comprehensive hardware designing procedures including silicone wings fabrication steps, mechanical crank slider mechanism details and integral backstepping based quadrotor control system with residual learning based wing controller from human demonstration.

We also show that our proposed method was able to achieve better deceleration performance compared to nominal quadrotor system. This results give us interesting future research direction for our hardware system including following sharp acceleration trajectories and exploration on confined complex indoor environments.

\addtolength{\textheight}{-5cm}   



\section*{APPENDIX}

Training of learning is done in NVIDIA RTX2060 with personal labtop.

\subsection{Hyperparameter for Supervised training}
We used same transformer encoder, decoder structure from \cite{petrovich2021action} with modified model size.
Our latent dimension of transformer network is $z \in \mathbb{R}^{32}$, feedforward networks size of 128, number of head of 8, number of layer of 2, with batchsize of 101. All the activation function was set be Rectified Linear Unit(ReLU).
States are in $\mathbb{R}^{12}$ and output action is in $\mathbb{R}^{1}$. Learning rate of training is 1e-5, KL divergence regularization term was 1e-6, and trained for 1500 epoch.

\subsection{Hyperparameter for RL training}
We created custom quadrotor training environments in isaac gym simulator \cite{makoviychuk2021isaac}. Our policy network and value network is parameterized by Multi Layer Perceptron(MLP) network with hidden layer size of [128, 64, 32]. We trained with 0.05s data with 8192 batch size. Hyperparameters related to PPO is set to default.


\bibliographystyle{IEEEtran}
\bibliography{ref.bib}

\end{document}